\documentclass{article}

\usepackage{float}
\usepackage{amsmath}
\usepackage{graphicx}
\usepackage{booktabs}
\usepackage{subcaption}
\usepackage{amsmath} 
\usepackage{amsfonts}
\usepackage{amssymb} 
\usepackage{pgffor}
\usepackage{enumitem}
\usepackage{pifont}

\DeclareUnicodeCharacter{2248}{$\approx$}

\usepackage[preprint]{corl_2026} 

\DeclareRobustCommand{\OURS}{MotionDisco: Motion Discovery for Extreme Humanoid Loco-Manipulation}

\title{\OURS}

\author{
  Ilyass Taouil$^{1,*}$ \And
  Michal Ciebielski$^{1,*}$ \And
  Shafeef Omar$^{1,*}$ \And
  Haizhou Zhao$^{1,2}$ \And
  Angela Dai$^{1}$ \And
  Aaron M. Johnson$^{3}$ \And
  Majid Khadiv$^{1}$ \And
  \normalfont\small $^{1}$Technical University of Munich, Germany \quad
  $^{2}$New York University, USA \quad
  $^{3}$Carnegie Mellon University, USA \\
  \small $^{*}$Equal contribution
}

\begin{document}
\maketitle
\begin{figure}[H]
  \centering
  \includegraphics[width=0.95\textwidth]{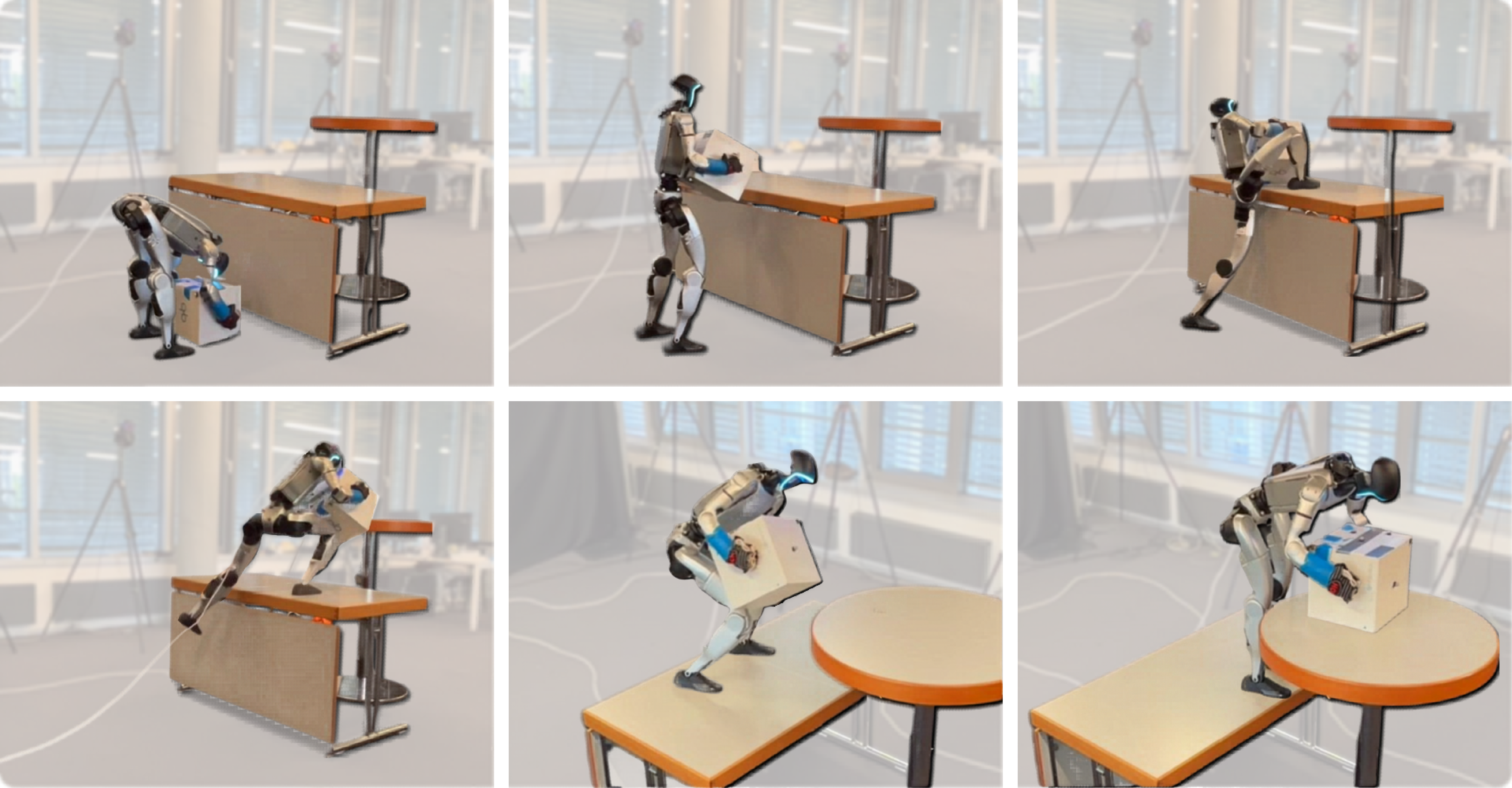}
    \caption{\textbf{Real-world humanoid motion from MotionDisco.} Snapshots of a real-world experiment executing a trajectory discovered by our evolutionary tree search, deployed zero-shot on the robot.}
  \label{fig:teaser}
\end{figure}

\begin{abstract}
We present \textit{MotionDisco}, a framework that discovers contact-rich, long-horizon humanoid loco-manipulation motions from scratch, without relying on teleoperation or motion retargeting from human demonstrations. This is challenging because the space of possible contact interactions grows combinatorially with the task horizon and the number of objects in the scene. MotionDisco enables rapid discovery of novel motions by coupling a large language model (LLM) guided evolutionary search over sequences of interactions with an efficient sequential kinodynamic trajectory optimizer and pruning strategy, enabling the rapid discovery of novel skills. Through extensive ablation studies, we show that our LLM-guided search discovers successful whole-body trajectories across several challenging long-horizon tasks. Finally, by training reinforcement learning tracking policies on the discovered trajectories, we transfer the motions to a real humanoid robot. This is the first work to discover and deploy long-horizon humanoid loco-manipulation skills entirely through automated evolutionary search. Supplementary videos of the experiments are available at: \url{https://youtu.be/DHiVz34QYlw}.
\end{abstract}
\keywords{Motion Discovery, Humanoid Loco-Manipulation, Program Search} 


\section{Introduction}
\label{sec:introduction}
For humanoid robots to operate effectively in the real world, they must be capable of executing long-horizon loco-manipulation behaviors in novel and unstructured environments. However, exploration in such high-dimensional, contact-rich settings remains extremely challenging \cite{tonneau2018efficient,toussaint2018differentiable,ciebielski2025task}. As a result, most state-of-the-art approaches rely either on retargeting human motion to humanoids to make exploration tractable or on teleoperation to bypass the exploration problem altogether. Despite their success, motion-retargeting approaches have inherent limitations, most notably the impracticality of generating demonstrations for every possible scenario a robot may encounter.

In this work, we move beyond pure imitation and propose a motion discovery framework that enables humanoid robots to reason about their environment and synthesize novel long-horizon behaviors. Specifically, we combine the reasoning capabilities of large language models (LLMs) with evolutionary algorithms to efficiently explore the space of contact interactions via program search, in which an LLM iteratively mutates code that generates candidate contact plans. Our framework is built around a layered kinodynamic trajectory optimization formulation that takes candidate contact sequences as input and progressively solves increasingly complex optimization problems. The resulting optimization feedback is used as numerical and language-based guidance for the program search process. Through extensive ablation studies and experimental evaluations, we demonstrate that the proposed framework can unlock highly complex humanoid loco-manipulation behaviors. 

The main contributions of this work are as follows:
\begin{itemize}[noitemsep, topsep=-\parskip]
    \item A framework that couples LLM-guided evolutionary search over contact plans with a kinodynamic trajectory optimizer that returns numerical and structured language feedback, enabling time-efficient discovery of complex long-horizon loco-manipulation behaviors.
    
    \item Extensive analysis showing that our search-based approach outperforms a non-iterative approach, indicating that search with structured feedback is essential for successful motion discovery. Moreover, the evolutionary nature of the search yields diverse motions for the same task in a single run, providing an efficient source of data generation for humanoid loco-manipulation.
    
    \item Real-world results on a wide range of long-horizon loco-manipulation behaviors. This is the first work to autonomously discover and execute long-horizon loco-manipulation behaviors on real hardware, without any motion retargeting from human data or teleoperation.
\end{itemize}
\section{Related Work}
\label{sec:related_works}

\subsection{Human to humanoid retargeting}
\label{sec:humand_to_humanoid_retargeting}

Retargeting human motion to humanoid robots has recently received significant attention. Human-to-humanoid motion transfer remains challenging due to morphological differences in degrees of freedom, limb lengths, and mass distributions. Existing methods~\cite{he2025asapaligningsimulationrealworld, he2024omnih2ouniversaldexteroushumantohumanoid} address this problem through keypoint-based inverse kinematics followed by unconstrained optimization to improve physical consistency. More recent approaches~\cite{omniretarget,taouil2025physicallyconsistenthumanoidlocomanipulation} formulate constrained trajectory optimization problems to improve the geometric consistency of the generated motions. 
To track these retargeted trajectories, several works employ reinforcement learning (RL) with reward formulations inspired by DeepMimic~\cite{peng2018deepmimic}, producing robust policies around reference motions and demonstrating successful tracking of retargeted human behaviors on humanoid hardware~\cite{beyondmimic}. More recent approaches extend this paradigm to loco-manipulation using proprioceptive observations, data augmentation, and contact-aware rewards~\cite{omniretarget, hdmi}.

While RL-based tracking can generate robust and dynamically feasible behaviors, it typically requires extensive training and high-quality demonstrations. In loco-manipulation settings, performance further depends on accurate contact information~\cite{resmimic, hdmi}, which is difficult to recover reliably from kinematic retargeting alone. Consequently, recent methods~\cite{pan2025spiderscalablephysicsinformeddexterous,dynaretarget} perform dynamic refinement of retargeted trajectories before RL tracking, leading to significant improvements. Despite recent progress in retargeting-based pipelines, they remain limited in complex or novel scenarios where relevant demonstrations are unavailable. Even when such data exists, tying motions to human demonstrations constrains the robot to human-like behaviors, preventing it from unlocking its full physical capability.

\begin{figure}
    \centering
    \includegraphics[width=1.0\textwidth]{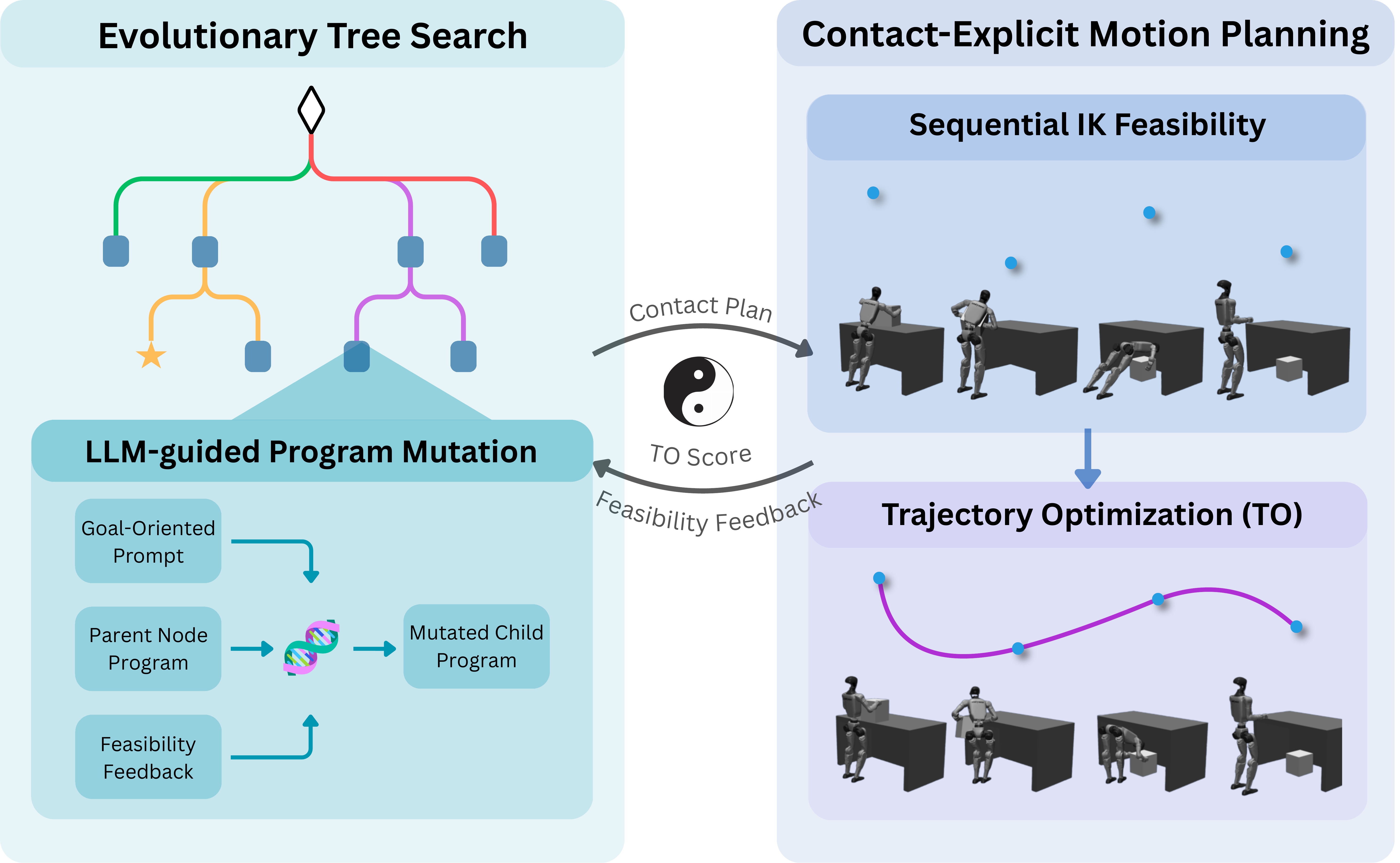}
    \vspace{1mm}
    \caption{\textbf{MotionDisco} couples LLM-guided evolutionary discovery of contact plans (left) with contact-explicit trajectory optimization (right). Each search node proposes a mutation of its parent program, conditioned on the goal prompt and the parent's feasibility feedback; executing the mutated program yields a discrete contact plan (denoted by the blue dots). Plans that pass a kinematic feasibility check are sent to the trajectory optimizer, which assesses dynamic feasibility and returns the feedback that guides subsequent mutations.}
    \label{fig:framework_overview}
\end{figure}

\subsection{LLMs for Motion Synthesis}
\label{sec:llms_for_motion_synthesis}

The common-sense reasoning capabilities of LLMs have recently sparked significant interest in their application to robotic motion planning. Early work by~\cite{ahn2022can} demonstrated how low-level robotic skills can be combined with LLMs that provide high-level procedural knowledge for executing complex, temporally extended instructions. Subsequently,~\cite{liang2023code} leveraged the code-generation capabilities of LLMs to translate natural language commands into robot policy code that processes perception outputs and parametrizes control primitives for tabletop manipulation tasks.

More recent approaches further integrate LLMs with planning and optimization. For example,~\cite{curtis2024trust} uses LLM-generated programs together with sampling or optimization methods to identify skill sequences that achieve task goals while satisfying constraints. Similarly,~\cite{chi2026instructflow} demonstrates the effectiveness of this strategy in a variety of long-horizon manipulation settings. In parallel,~\cite{ma2024eureka} employs LLMs within an evolutionary optimization loop over reward code, while~\cite{wu2025human} shows that hierarchical planning with LLMs enables more complex human behaviors that can subsequently be tracked using RL in simulation.

The work most closely related to ours is~\cite{shcherba2025meta}, which uses LLMs to perform program search in order to assist trajectory optimization for long-horizon tasks. In contrast to~\cite{shcherba2025meta}, which operates at the level of manipulation predicates, our approach directly generates programs over contact plans, enabling the treatment of both locomotion and manipulation behaviors. Furthermore, we introduce a trajectory optimization framework with progressively increasing structural complexity that provides informative feedback to the program search process. In particular, we build on recent advances in program evolution and scientific discovery~\cite{novikov2025alphaevolve,lange2025shinkaevolve,cemri2026adaevolve,liu2026evox} to generate candidate contact plans, which we evaluate with an efficient kinodynamic trajectory optimization tool to drive a sample-efficient evolutionary search over contact sequences.


\section{Method}
\label{sec:method}

This section presents MotionDisco, a framework for discovering long-horizon humanoid loco-manipulation behaviors through the interaction of discrete contact-plan search and continuous motion planning. The framework alternates between proposing candidate contact sequences, which explore the combinatorial space of possible interactions, and evaluating whether each candidate can be realized as a feasible robot--object motion consistent with the scene geometry and dynamics. We first describe the contact-explicit motion-planning formulation used to evaluate candidates and generate feasible trajectories, and then the LLM-guided evolutionary search, which represents contact plans as programs, mutates them through language-model-guided edits, and uses motion-planner feedback to improve subsequent proposals. Figure~\ref{fig:framework_overview} illustrates the overall framework.

\subsection{Motion Planning}
\label{sec:motion_planning}

The motion planning module evaluates the contact plans produced by the LLM-guided search and is used in two ways. First, it acts as an evaluator for contact-mode discovery: a sequential kinematic optimization, similar to~\cite{ciebielski2026discovery}, rejects contact sequences that fail to satisfy collision avoidance, configuration limits, and transition consistency, returning text feedback that identifies where the proposed interaction sequence fails. This stage ignores dynamics and serves as a fast geometric feasibility filter. Second, when a sequence is feasible, the planner solves for the robot--object state trajectory, controls, contact wrenches, and stage durations while enforcing dynamics, contact constraints, collision avoidance, and state/control limits; the resulting trajectory is used for downstream policy training and execution. 

Let $\mathcal{I}$ be the set of contact interfaces.
For each $a \in \mathcal{I}$, define its contact state as a pair $(a,\, b)$ where $b \in \mathcal{I}$ denotes unilateral contact with interface $b$, or $(a, \varnothing)$ if free. The interfaces $a$ represent the contact interfaces that can change their state, such as end effectors and movable objects. A contact mode is the complete assignment over all interfaces:
\begin{equation}
    c = \{(a,\, b) \forall a \in \mathcal{I}: \ b \in \mathcal{I} \cup \{\varnothing\}\},
\end{equation}
with a consistency constraint that if $b\neq \varnothing$ then $(a,b) \in c \Rightarrow (b,a) \in c$.
For a contact mode $c$, let $\mathcal{M}(c)$ denote the feasible constraint manifold associated with it:
\begin{equation}
\label{eq:conconstrman}
    \mathcal{M}(c) :=
    \left\{
    (x,\dot x,u)\;\middle|\;
    \begin{aligned}
        &\mathrm{dynamics}(x,\dot x,u,c)=0,\\
        &\mathrm{contact}(x,u,c)\leq 0,\\
        &\mathrm{collision}(x)\leq 0,\\
        &\mathrm{limits}(x,u)\leq 0
    \end{aligned}
    \right\}
\end{equation}
The set $\mathcal{M}(c)$ contains all state $x,\dot x$, and control $u$ tuples that satisfy the dynamics and contact conditions induced by $c$, remain within hardware limits, and avoid unintended collisions. The latter requirement is particularly important, since any additional collision would create an unmodeled contact and thus imply a departure from the contact mode $c$.

Given a fixed contact-mode sequence $\mathcal{C}=(c_0,\ldots,c_{K-1})$, the continuous optimization problem finds the state trajectory $x(\cdot)$, control trajectory $u(\cdot)$, and stage durations $\bar T_{0:K-1}$, where each $\bar T_k$ is the duration of mode $c_k$. These durations induce switching times $t_0=0$, $t_{k+1}=t_k+\bar T_k$, and final time $T=t_K$, with $c(t)=c_k$ for $t\in[t_k,t_{k+1})$. The continuous trajectory optimization problem induced by a fixed contact-mode sequence is defined as follows:
\begin{equation}
\label{eq:optimization_formulation}
    \begin{aligned}
    \min_{\substack{x(\cdot),\,u(\cdot),\,\bar T_{0:K-1}}}
    & \int_0^T \phi(x(t),u(t))\,dt
      + \phi_T(x(T))
      + w_T T \\
    \text{s.t.}\quad
    & x(0)=x_{\mathrm{init}}, \quad
      c_0=c_{\mathrm{init}}, \quad x(T)\in X_{\mathrm{goal}},\\
    & \forall k \in \{0,\dots,K-1\}:\\
    & \quad \bar T_{\min}\leq \bar T_k \leq \bar T_{\max},
     \quad (x(t),\dot x(t),u(t)) \in \mathcal{M}(c_k),
      \quad \forall t \in [t_k,t_{k+1}].
    \end{aligned}
\end{equation}
Where, $\phi$ is the running cost, $\phi_T$ is the terminal cost, $w_T$ weights the final-time penalty, and $X_{\mathrm{goal}}$ is the desired goal set. 

Both the sequential kinematic optimization and the trajectory optimization are implemented using a direct multiple shooting transcription and solved with temporal structure exploiting solvers, \cite{Verschueren2021acados} and \cite{zhao2026hippo}, respectively.

\subsection{Motion Discovery via LLM-guided Evolutionary Search}
\label{sec:llm_planner}
Given a scene representation $\mathcal S$ and a goal $\mathcal G$, our objective is to discover a contact-mode sequence
\(
\mathcal C=(c_0,\ldots,c_{K-1})
\)
that reaches the goal and respects all constraints.
The scene representation $\mathcal S$ is a text-based description of the world containing the label, position, and dimensions of every object in the scene. The goal $\mathcal G$ may contain both discrete and continuous components: the discrete component specifies desired contacts between robot or object interfaces, while the continuous component specifies target states such as robot or object poses.

The remainder of this section describes the full search loop. We first define the tree search representation, in which each node stores an executable contact-plan program together with its score and feedback. We then describe the parent-selection policy used to choose nodes for expansion, the LLM-based mutation operators used to generate new contact-mode sequences, and the automatic scoring and feedback procedure that evaluates each candidate through kinematic feasibility and trajectory optimization. All LLM-based components in this work use \textit{Claude Opus 4.7}~\cite{anthropic2026claude47}.

\subsubsection{LLM-guided Tree Search}
\label{sec:tree_search}

We search over contact-mode sequences using an LLM-guided evolutionary tree
$T=(N,E)$. Each node $n\in N$ is a self-contained candidate contact plan represented by a program $p_n$ whose execution returns a sequence $\mathcal C_n$. The node also stores a scalar score $F_n$, optional text feedback $\tau_n$, and an offspring count $o_n$ recording how many children have already been generated from that node. Each edge $e\in E$ corresponds to a mutation from a parent program to a child program. Starting from a root node encoding the initial contact state, the search iteratively selects a parent node, mutates its contact-plan program with an LLM, evaluates the resulting sequence, and adds the evaluated candidate back to the tree. The search terminates when the allocated search budget is exhausted.

\subsubsection{Tree Expansion}
\label{sec:tree_expansion}

LLM-guided program evolution methods~\cite{novikov2025alphaevolve,lange2025shinkaevolve,cemri2026adaevolve,liu2026evox} rely on several mechanisms for effectively expanding the tree search. First, \textit{expansion sampling} determines which existing nodes are selected for further expansion. Second, \textit{diversity maintenance} prevents premature convergence by encouraging the search to preserve multiple promising directions. Third, \textit{inspiration sampling} selects additional nodes from the tree to include in the prompt, allowing new candidates to build on diverse prior solutions. In this work, we adopt the tree expansion strategy of ShinkaEvolve~\cite{lange2025shinkaevolve}, which additionally uses novelty-based rejection sampling to filter candidates too similar to existing ones.

In ShinkaEvolve, these mechanisms are realized through a hierarchical population model. Expansion sampling is island-conditioned: candidate parents are sampled from structured subpopulations, which distributes search effort across different regions of the tree. The same island structure provides diversity maintenance by preserving multiple subpopulations and preventing the search from collapsing around a small set of high-scoring nodes. For prompt construction, ShinkaEvolve samples additional tree nodes as inspirations, allowing the LLM to condition candidate generation on diverse prior solutions. The details of each of these steps can be found in Appendix~\ref{sec:expansion_details}.

\subsubsection{Contact Plan Mutation}
\label{sec:contact_plan_mutation}

After selecting a parent node, MotionDisco constructs an LLM mutation prompt from the scene $\mathcal S$, goal $\mathcal G$, parent program $p_n$, parent score $F_n$, and parent feedback $\tau_n$. The prompt also specifies the valid interfaces $\mathcal I$, so that mutations are restricted to meaningful contact assignments in the current scene. Details about the prompt can be found in Appendix~\ref{sec:prompt_details}

The LLM returns a modified program $p'$, whose execution produces a new contact-mode sequence:
\begin{align}
\mathcal C'=(c'_0,\ldots,c'_{K'-1}),
\qquad
c'_s=\{(a,b'):a\in\mathcal I\}.
\end{align}
Thus, the LLM mutates only the discrete contact plan: each stage must assign exactly one target ${b}\in\mathcal I\cup\{\varnothing\}$ to every active interface $a$, while the continuous motion between contact switches is recovered later by kinematic feasibility and trajectory optimization.

The prompt conditions each mutation on the current tree state through the selected parent and, optionally, other nodes sampled from the tree. After parsing, the proposed program's contact plan is checked for syntactic validity and consistency with the contact-mode definition in Sec.~\ref{sec:motion_planning}. Valid proposals are evaluated by the motion planner and inserted into the tree as children of the selected parent.

\subsubsection{Contact Plan Scoring and Feedback}
\label{sec:contact_plan_scoring}

Given a candidate sequence, we first run the sequential kinematic feasibility check; if the full sequence is kinematically feasible, we then solve the trajectory optimization problem. In this case, the cost assigned to the plan is computed from the resulting trajectory as the sum of the constraint residuals and a jerk regularization term:
\begin{align}
J(\mathcal C') =
\sum_i \left\|r(x_i,u_i,c_{s(i)})\right\|_1
+
w_j \sum_i \left\|\dddot q_i\right\|^2 ,
\end{align}
where $r(x_i,u_i,c_{s(i)})$ collects the equality and inequality constraint residuals at knot point $i$, and $w_j$ weights the jerk penalty. The text feedback reports if the full contact sequence is kinematically feasible together with the resulting residual and jerk costs.

If the full kinematic sequence is infeasible, trajectory optimization is not run and a score of zero is given to the respective node. Instead, we identify the failure point by iteratively shortening the sequence until the longest feasible prefix is found:
\begin{align}
\hat K
=
\max\left\{
k\in\{0,\ldots,K'\}:
(c'_0,\ldots,c'_{k-1}) \text{ is kinematically feasible}
\right\}.
\label{eq:kine_feasible}
\end{align}
The first failing mode or transition, namely $c'_{\hat K}$ or $c'_{\hat K-1}\rightarrow c'_{\hat K}$, is returned as text feedback to the LLM. This feedback describes which contact assignment caused the kinematic optimization to fail, allowing future mutations to repair the contact plan locally rather than discard the entire sequence.

\section{Evaluation}
\label{sec:evaluation}

We evaluate MotionDisco on eight tasks spanning a range of difficulty levels, from simple pick-and-place to complex whole-body behaviors that require integrating locomotion, manipulation, and reasoning about the environment as a source of support. Detailed task descriptions and scene illustrations are provided in Appendix~\ref{sec:tasks}. We organize our evaluation in three parts. First, we quantitatively analyze motion discovery across different expansion and feedback strategies, reporting the best trajectory optimization cost achieved, the percentage of valid contact plans found, and the time required to find the first valid solution for the best strategy. Second, we qualitatively analyze the diversity of motions discovered by the search on a representative scene, showing that it finds a variety of contact plans that all satisfy the goal in a single search. Finally, we carry out extensive real-world experiments showing that the discovered motions transfer successfully to the real humanoid robot.

\subsection{LLM-guided Search Enables Motion Discovery}
\label{sec:ablation_1}

Table~\ref{tab:ablation_costs} reports the cost of the recovered motions under three regimes, namely a single LLM call (SC), MotionDisco without text feedback (MD w/o TF), and our evolutionary search framework MotionDisco with text feedback (MD). The text feedback consists of a textual description informing the LLM at which node the contact plan becomes kinematically infeasible, \eqref{eq:kine_feasible}. Unlike SC, both MD w/o TF and MD successfully solve all scenarios, highlighting that a single LLM query is insufficient and that iterative search is essential to refine the initial proposals and discover feasible, lower-cost contact plans. Moreover, given the same search budget, MD finds a higher percentage of valid contact plans at a lower TO cost, demonstrating the benefit of text feedback in guiding the search. Importantly, MD finds a first valid solution within a few minutes across all scenarios, making the approach practical despite the combinatorial size of the contact-plan space. Here, we consider a contact plan valid if it first passes the sequential kinematic feasibility check and the TO converges. 
\begin{table}[t]
    \centering
    \caption{Comparison across different expansion and feedback strategies. We additionally report the time required to find the first valid solution for the best strategy (MotionDisco, MD).}
    \label{tab:ablation_costs}
    \setlength{\tabcolsep}{4pt}
    \begin{tabular}{r l c c c c c c}
        \toprule
        & & \textbf{SC} & \multicolumn{2}{c}{\textbf{MD w/o TF }} & \multicolumn{3}{c}{\textbf{MD}} \\
        \cmidrule(lr){3-3} \cmidrule(lr){4-5} \cmidrule(lr){6-8}
        \textbf{\#} & \textbf{Task} & \textbf{Cost} $\downarrow$ & \textbf{Cost} $\downarrow$ & \textbf{Valid \%} $\uparrow$ & \textbf{Cost} $\downarrow$ & \textbf{Valid \%} $\uparrow$ & \textbf{Time} [min] $\downarrow$ \\
        \midrule
        1 & Banana                    & /      & 21.17 & 48.14 & \textbf{20.08} & \textbf{70.30} & 7.49 \\
        2 & Box Stacking              & 398.05 & 398.05 & 45.45 & \textbf{10.53} & \textbf{63.63} & 4.58 \\
        3 & Climb Table w/ Box        & 16.16  & 16.47 & 46.15  & \textbf{15.97} & \textbf{53.84} & 2.15 \\
        4 & Long-Dist.\ Pick \& Place & 13.61  & 10.70 & 65.00 & \textbf{10.42} & \textbf{70.00} & 2.50 \\
        5 & Move Through Clutter      & /      & 11.96 & \textbf{73.33} & \textbf{11.67} & \textbf{73.33} & 3.11 \\
        6 & Parkour Pick \& Place 1   & /      & 15.97 & 33.33 &  \textbf{15.47}   & \textbf{38.88} & 6.05 \\
        7 & Parkour Pick \& Place 2   & /      & 15.90 & 58.82 & \textbf{15.51} & \textbf{67.64} & 4.40 \\
        8 & Under-Table Pick \& Place & 115.64 & 18.87 & 64.28  & \textbf{17.84} & \textbf{71.42} & 1.34 \\
        \bottomrule
    \end{tabular}
\end{table}
\subsection{LLM-guided Search Produces Diverse Motions}
\label{sec:qualitative_2}

\begin{figure}[t]
    \centering
    \begin{subfigure}{\textwidth}
        \centering
        \includegraphics[width=0.9\textwidth]{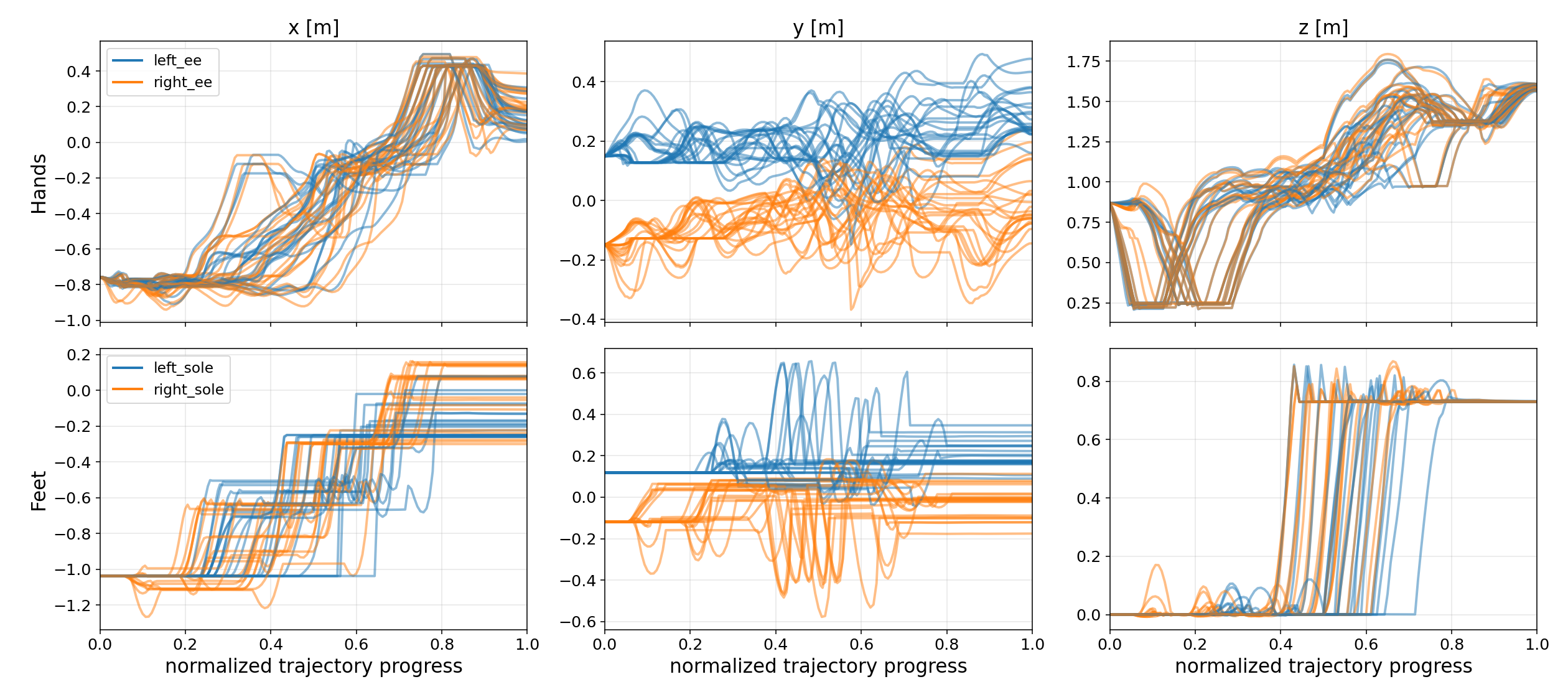}
        \label{fig:end_effector_trajectories}
    \end{subfigure}
    \\[1ex]
    \begin{subfigure}{\textwidth}
        \centering
        \includegraphics[width=0.9\textwidth]{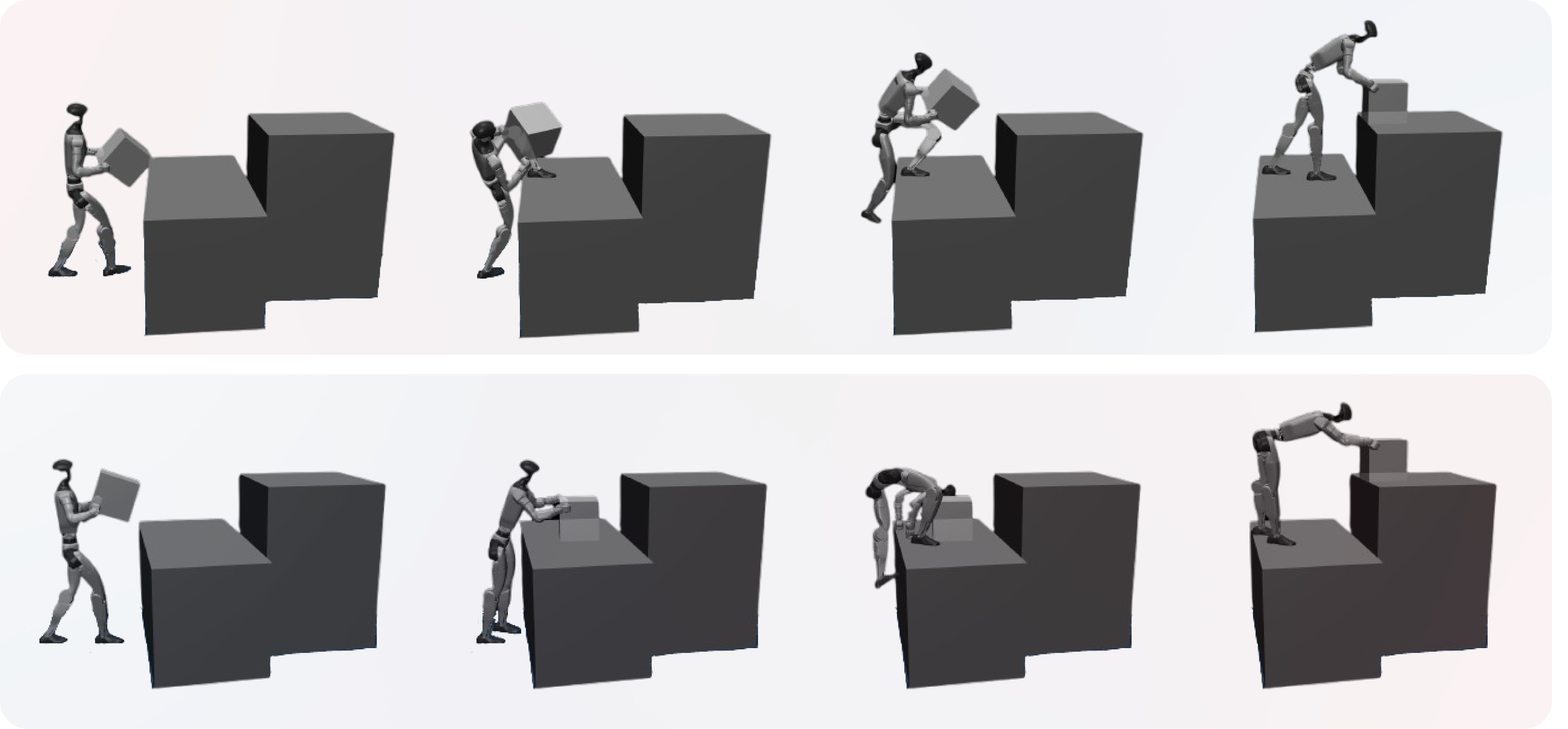}
        \label{fig:whole_body_trajectory}
    \end{subfigure}
    \caption{\textbf{Motion diversity discovered by the search on Parkour Pick \& Place 2.} \textit{Top:} $xyz$ hand and foot trajectories across all valid contact plans found by MD. \textit{Bottom:} whole-body trajectory snapshots of two distinct solutions. The variation in both panels arises from the diversity of the underlying contact plans found in a single run.}
    \label{fig:diversity_combined}
\end{figure}

Beyond cost reduction, the evolutionary search naturally produces multiple distinct solutions for the same task. Because the population preserves diversity across generations and duplicate plans are penalized, the search returns qualitatively different contact plans that all satisfy the goal—for example, alternative assignments of end-effectors to environment features. These variations emerge without re-prompting or restarting the pipeline, providing a source of motion diversity that a single LLM call—which tends to collapse onto a single mode—cannot offer. Figure~\ref{fig:diversity_combined} illustrates the motion diversity for the \textit{Parkour Pick \& Place 2} task.
\subsection{Real-World Deployment}
\label{sec:real_world}

To validate MotionDisco on hardware, we train motion tracking policies for the discovered trajectories using RL with DeepMimic-style rewards~\cite{peng2018deepmimic} and domain randomization, and execute them zero-shot on a real humanoid robot. The resulting controllers robustly reproduce the discovered motions, achieving multiple consecutive successful executions across all attempted tasks. Figure~\ref{fig:teaser} shows snapshots of one such execution, with additional snapshots in Appendix~\ref{sec:real_world_deployment}. We refer the reader to the supplementary video for further results.

\section{Conclusion}
\label{sec:conclusion}

This paper presents MotionDisco, a framework for discovering long-horizon whole-body humanoid motions. MotionDisco couples an LLM-guided evolutionary search over discrete contact plans with a layered kinodynamic motion-planning module, in which a sequential kinematic feasibility stage rapidly prunes geometrically inconsistent plans and a full trajectory optimization stage produces dynamically feasible robot--object motions for the surviving candidates. The structured failure feedback returned by the motion planner is translated into language and used to guide subsequent LLM mutations, closing the loop between high-level reasoning and low-level optimization. Across eight long-horizon loco-manipulation tasks, our search consistently outperforms a single-call state-of-the-art LLM in both success and trajectory cost, and naturally produces qualitatively different solutions to the same task. Finally, through real-world experiments, we demonstrate the first humanoid loco-manipulation behaviors of this complexity generated without motion retargeting or teleoperation.

\section{Limitations \& Future Work}
\label{sec:limitations_and_future_work}

While MotionDisco enables the discovery of complex loco-manipulation behaviors, several limitations remain. First, the sequential kinematic and trajectory optimization used in MotionDisco currently supports only unilateral sticking contacts on rectangular patches, and the manipulated objects are restricted to rigid box-like geometries. Extending the contact model to sliding and non-unilateral contacts, and the object set to articulated and free-form shapes, would significantly broaden the range of behaviors and environments the framework can address. Second, MotionDisco assumes a known scene and does not currently incorporate perception; coupling the search with a vision module that builds the scene representation directly from sensor observations would be an important step toward deployment in unstructured environments. Finally, another important next step to improve MotionDisco is to integrate the recent advances in simulation-ready scene construction for on-the-fly motion discovery \cite{xia2026simreconsimreadycompositionalscene,xia2026holoscene,dong2026replicateanyscene}.

\clearpage


\bibliography{example}  

\newpage
\appendix

\section{LLM-guided Search Implementation}
\label{sec:llm_guided_search}

\subsection{Prompt Details}
\label{sec:prompt_details}

The prompt provides the LLM with task-specific scene information, contact-surface identifiers, planning rules, helper functions, and evaluator feedback, then asks it to generate a contact mode sequence given a goal description. It also includes examples of useful contact-transition patterns, such as two-hand grasps, lifts, placements, releases, and step-ups. These patterns are intended as illustrative guidance rather than required templates: the model may use, combine, or depart from them as needed, provided the resulting contact plan satisfies the task constraints and contact validity rules.

{\footnotesize
\begin{verbatim}
============================================================
You are a contact planner for a humanoid robot. Given a task
and a scene with labeled surface IDs, you output Python code that builds a
contact_mode_seq — a dictionary describing how the robot's end-effectors, soles,
and manipulated objects make and break contact across a sequence of modes.

Your job is to reason about the INTERESTING subgoals only: grasps, releases,
lifts, placements, and step-ups onto elevated surfaces. Ordinary walking
between subgoals is delegated to a standard locomotion helper, walk(), so you
just decide how many steps to take and never have to hand-craft plain stepping
modes.

# Notation
# EEs and soles:   (env_id, contact_type)          or () for free
# Objects:         (self_id, env_id, contact_type) or () for free
#
# env_id:          ID of the surface in the scene being contacted
# self_id:         ID of the surface on the object itself
# contact_type:    1 = unilateral (only supported value)
#
# A mode is a full snapshot of all active contacts. Each mode lasts 1 second.
# Only emit a new mode when at least one contact actually changes.

# Rules
- At least one sole must be in contact at all times.
- Object channels are passive: env contact breaks when lifted, and only
  reappears when the object is explicitly placed or lands on a new surface.
  When placed on another object, env_id becomes that object's surface ID.
- Concurrent make/break events in one transition are intentional and encouraged.
- The robot has flat palms with no finger closure. To hold an object
  against gravity, BOTH EEs must be in contact with it — a single EE
  cannot grip an object alone. Both EEs must be in contact before the
  lift transition (object channel -> ()), and must remain in contact
  until the object is placed back on a support surface. The two EEs
  may make contact in different modes (sequential grasp), but both
  must be on the object before the object becomes free.
- Be efficient: combine concurrent events into a single mode, no noop repeats,
  no modes that change nothing. Shorter plans solve faster and rarely score
  worse than padded ones.

# Helpers available in scope

    get_initial_mode() -> Dict[str, List[Tuple]]
        Returns the scene's compiled starting mode as a length-1
        contact_mode_seq dict. Use it as mode 0 of your plan, then
        grow each channel by calling walk() and append_mode():
            plan = {ch: list(seq) for ch, seq in get_initial_mode().items()}
        The list(seq) copy is required so each channel is a mutable
        list the helpers can append to.

    walk(plan, n_steps) -> None
        Appends a standard locomotion pattern of n_steps steps to `plan`.
        One step = one foot lifts and plants again (alternating feet, right
        foot first). The feet stay on the support surface they were already
        on. All non-sole channels are held at their current values
        throughout the walk, so any active EE grasp and attached object
        persist through it.

        walk produces a STRAIGHT-LINE base translation toward the next
        contact-implied target — it does NOT plan around obstacles. If any
        movable or fixed object's xy footprint intersects the line segment
        from the current base xy to the next required base xy, the walk
        will collide. You must clear movable obstacles (grasp, lift, walk
        aside, place) BEFORE the walk that would pass through them. Fixed
        obstacles must be routed around explicitly — typically with an
        intermediate walk to a waypoint that gives a clear line.

        After walk returns the plan is in DOUBLE SUPPORT: both soles are
        planted on their original support env_ids (only the base xy has
        moved). This holds for any n_steps. A step-up, step-down, or any
        other foot transition chained immediately after walk therefore
        still begins with a foot-lift (sole -> ()) mode.

        walk does NOT step onto elevated surfaces (box lids, platforms) or
        change support surfaces. Use append_mode for those transitions.

    append_mode(plan, **channel_updates) -> None
        Appends exactly one new mode to `plan`. Any channel not mentioned
        keeps its previous value. Use this to express interesting transitions.

# When to use walk vs append_mode

Use walk whenever the only thing changing between two subgoals is the base's
xy position on flat ground. Don't enumerate the stepping modes yourself —
just decide how many steps the robot needs and let IK/TO place them.

Use append_mode for everything that actually matters for the task:
  - grasping:     EE goes from ()             -> (env_id, 1)
  - releasing:    EE goes from (env_id, 1)    -> ()
  - lifting:      object goes from (self_id, env_id, 1)
  - placing:      object goes from ()         -> (self_id, env_id, 1)
  - stepping up / down onto a different support surface

# Interesting patterns

In the patterns below, env_id_1, env_id_2, ... are placeholders for distinct
env_ids drawn from the active scene's ID table (similarly self_id_1 for
self_ids). Within a single pattern the same name refers to the same id;
across patterns the names are independent. Substitute the actual integers
from the ID table when emitting your plan.

The patterns are illustrative, not exhaustive. Channels are independent
and any combination of transitions consistent with the rules above is
valid — for instance, EEs may press against environment surfaces (not
just object surfaces) to provide support during a step-up or step-down.

## Two-EE grasp + lift, concurrent
append_mode(plan, left_ee=(env_id_1, 1), right_ee=(env_id_2, 1))
append_mode(plan, box1=())

## Two-EE grasp + lift, sequential (stabilize, then secure)
append_mode(plan, left_ee=(env_id_1, 1))
append_mode(plan, right_ee=(env_id_2, 1))
append_mode(plan, box1=())

## Place and release
append_mode(plan, box1=(self_id_1, env_id_1, 1))
append_mode(plan, left_ee=(), right_ee=())

## Step up onto an elevated surface
append_mode(plan, right_sole=())
append_mode(plan, right_sole=(env_id_1, 1))
append_mode(plan, left_sole=())
append_mode(plan, left_sole=(env_id_1, 1))

## Step up onto an elevated surface, assisted by one EE
# One EE presses against the destination surface for support; the EE is
# established before the feet move and released once both feet are planted.
append_mode(plan, left_ee=(env_id_1, 1))
append_mode(plan, right_sole=())
append_mode(plan, right_sole=(env_id_1, 1))
append_mode(plan, left_sole=())
append_mode(plan, left_sole=(env_id_1, 1))
append_mode(plan, left_ee=())

## Step up while carrying an object
# Same as above — EE and object channels inherit, so the grasp persists.
append_mode(plan, right_sole=())
append_mode(plan, right_sole=(env_id_1, 1))
append_mode(plan, left_sole=())
append_mode(plan, left_sole=(env_id_1, 1))

## Step down from an elevated surface back to the floor
# Currently both feet on an elevated surface, stepping back to the floor.
append_mode(plan, right_sole=())
append_mode(plan, right_sole=(env_id_1, 1))
append_mode(plan, left_sole=())
append_mode(plan, left_sole=(env_id_1, 1))

## Step down from an elevated surface, assisted by one EE
# Currently both feet on env_id_1 (elevated). Descending to env_id_2 (lower).
# One EE presses against the surface being left for support during descent.
append_mode(plan, left_ee=(env_id_1, 1))
append_mode(plan, right_sole=())
append_mode(plan, right_sole=(env_id_2, 1))
append_mode(plan, left_sole=())
append_mode(plan, left_sole=(env_id_2, 1))
append_mode(plan, left_ee=())

# Geometric reasoning
- Before writing a grasp, place, or step-up, check the target's xyz in the
  scene geometry block and compare it to where the base will be at that
  point in the plan. The G1 can reach ~0.8 m from the base; anything beyond
  could require walking or climbing to be within reach.
- Before EVERY walk, mentally trace the straight line from the current base
  xy to the next required base xy (implied by the next subgoal). If that
  line passes through any object's xy footprint — movable OR fixed — the
  walk will collide and IK will fail at that mode. Movable objects in
  the way MUST be relocated before the walk; fixed objects (walls,
  pillars) MUST be routed around. There is no other way through; "just
  walk past it" is not an option the solver can rescue.
- The robot body has non-zero width (~0.4 m). Treat any object whose
  footprint comes within roughly 0.2 m of the straight-line path as in
  the way, not just objects exactly on the line.
- Estimate step count from distance: one G1 step covers ~0.4 m, so a 1 m
  approach is ~3 steps, a 2 m approach is ~5 steps, a 3 m approach is
  ~8 steps. Don't over-step — extra steps inflate the expanded plan
  length and burn the mode budget. Round to the nearest integer; only
  round up if you'd otherwise end the walk out of arm's reach for the
  next subgoal.
- Track the base mode-by-mode. Soles move the base: xy follows the feet,
  z ≈ sole-surface z + ~0.8 m. Stepping onto a box (via append_mode)
  raises the base by roughly the box's size_z. EE and object operations
  do NOT move the base.
- Step-up height limit: the robot can only step onto a surface that is at
  most 0.8 m above its current support surface. Compute the height
  difference as (target_surface_z) - (current_support_surface_z); if it
  exceeds 0.8 m the step-up is infeasible and IK will fail. To reach a
  higher surface, stage the climb via an intermediate surface or 
  hand assistend climbing.
  The same limit applies symmetrically to step-downs.
- walk does not change which surface the feet are on. To get onto an
  elevated surface, emit explicit step-up modes via append_mode.
- Re-read your plan before emitting it: at every append_mode, verify that
  the new contact is reachable from the base position implied by everything
  before it. If not, insert a walk (or step-up sequence) first.

# Planning order
Before writing any code, do this audit in order:
  1. Identify the final subgoal(s) the task requires.
  2. List every movable object whose xy footprint lies between the robot's
     start position and any point the base must visit. These are obstacles
     and need to be relocated first, even if the task description doesn't
     mention them.
  3. Write the plan: clearing subgoals first, then task subgoals.

# Output format
First state the high-level plan in one line, e.g.:
    "walk to box, two-EE grasp, lift, walk to table, place on table, release"
Then emit the Python body of generate_contact_plan(scene). It must start from
get_initial_mode() and return the resulting dict.

# Feedback
If the IK solver finds the expanded plan infeasible, the evaluator will report
at which mode feasibility broke. Mode indices start at 0, where mode 0 is the
scene's initial state, and refer to the FULLY EXPANDED plan. For example:

    "IK infeasible at mode 5 (of 8); mode 0 is the scene's initial state."

A failure inside a walk expansion usually means the surrounding grasp /
object state is incompatible with locomotion (e.g. carrying an oversized
object), or that the requested step count is wrong for the geometry — too
few steps leaves you out of reach for the next subgoal, too many runs you
into something. Adjust n_steps or rethink the subgoal order.

# ---- Active task: parkour_pick_place_2/default/place_box_on_table ----

Pick the white box and place it on the elevated table.

# Discrete contact goal
The FINAL mode of your contact_mode_seq must satisfy:
"white_box": final entry must be (6, 8, 1)  (self_id=6, env_id=8, contact_type=1)

Partial matches are rewarded inside the discrete-goal band — matching k of N 
channels gives goal progress k/N — but 
IK feasibility and TO cost are only evaluated 
when ALL channels match (a full pass is required to advance to the next band).

# Continuous cost term
These targets shape the trajectory: when IK passes and TO runs,
plans whose final state is closer to the targets converge with
lower TO cost (lower constraint violations + control jerk),
placing them higher inside the TO band of the score.

(goal: box_on_table)

  - object 'white_box' final xyz should be near (0.700, 0.000, 1.265)

# ---- Scene geometry ----

The scene's initial geometry (xyz in meters, size in meters):
  robot base: xyz=[-1.00, 0.00, 0.77]
  floor (fixed): xyz=[0.00, 0.00, 0.00] size=[1.5, 0.7]
  table_short (fixed): xyz=[0.00, 0.00, 0.36] size=[0.35, 0.7, 0.365]
  table_tall (fixed): xyz=[0.80, 0.00, 0.56] size=[0.395, 0.395, 0.565]
  white_box (movable): xyz=[-0.70, 0.00, 0.14] size=[0.1325, 0.1275, 0.135]

# ---- Active scene surface IDs ----

There are TWO distinct ID spaces in every contact tuple:

  env_id  (global feature index, used in EVERY tuple as the env_id)
  self_id (per-object feature index, used only in object channels)

For the active scene, the IDs introspected from the compiled
contact_tamp scene are:

  floor:
    env_id 1   = floor_feat1  (self_id 1)  [top]
  table_short:
    env_id 2   = table_short_feat1  (self_id 1)  [top]
    env_id 3   = table_short_feat2  (self_id 2)  [left]
    env_id 4   = table_short_feat3  (self_id 3)  [front]
    env_id 5   = table_short_feat4  (self_id 4)  [right]
    env_id 6   = table_short_feat5  (self_id 5)  [back]
    env_id 7   = table_short_feat6  (self_id 6)  [bottom]
  table_tall:
    env_id 8   = table_tall_feat1  (self_id 1)  [top]
    env_id 9   = table_tall_feat2  (self_id 2)  [left]
    env_id 10  = table_tall_feat3  (self_id 3)  [front]
    env_id 11  = table_tall_feat4  (self_id 4)  [right]
    env_id 12  = table_tall_feat5  (self_id 5)  [back]
    env_id 13  = table_tall_feat6  (self_id 6)  [bottom]
  white_box:
    env_id 14  = white_box_feat1  (self_id 1)  [top]
    env_id 15  = white_box_feat2  (self_id 2)  [left]
    env_id 16  = white_box_feat3  (self_id 3)  [front]
    env_id 17  = white_box_feat4  (self_id 4)  [right]
    env_id 18  = white_box_feat5  (self_id 5)  [back]
    env_id 19  = white_box_feat6  (self_id 6)  [bottom]

Each row is tagged with the feature's face direction in the object's
body frame (`top` / `bottom` / `front` / `back` / `left` / `right`).
Use it to pick the right self_id: `bottom` is what rests on a horizontal
surface, `top` is what receives a stack, opposing pairs (`front`/`back`
or `left`/`right`) are the natural two-EE grasp pair.

Required channels: "left_ee", "right_ee", "left_sole", "right_sole", "white_box"

EE/sole entries are (env_id, contact_type); object entries are
(self_id, env_id, contact_type); () always means "free".
contact_type = 1 (unilateral) is the only supported value for now.

The scene's compiled initial mode (use this as mode 0):

    contact_mode_seq = {
        'left_sole' : [(1, 1)],
        'right_sole': [(1, 1)],
        'left_ee'   : [()],
        'right_ee'  : [()],
        'white_box' : [(6, 1, 1)],
    }

Movable objects (each needs its own channel): "white_box"
============================================================
\end{verbatim}
}

\subsection{Tree Expansion and Sampling Details}
\label{sec:expansion_details}

\begin{itemize}
    \item \textbf{Expansion sampling (population model):} We use the hierarchical sampling policy from ShinkaEvolve~\cite{lange2025shinkaevolve}. The evaluated nodes are partitioned into island subpopulations. At each expansion step, an island is first sampled uniformly, and the parent node is then sampled from the evaluated nodes in that island. This separates global exploration across islands from local exploitation within an island.

    Let $\mathcal{I}$ denote the set of islands and let $N_{\mathrm{eval}}^{(k)}$ be the evaluated nodes
    in island $k \in \mathcal{I}$. We first sample
    \[
    k \sim \mathrm{Uniform}(\mathcal{I}).
    \]
    Within the selected island, each node $i \in N_{\mathrm{eval}}^{(k)}$ is assigned a weight based on its score and offspring count. Let $F_i$ denote the node score and $o_i$ the number of times node $i$ has already been expanded. We compute
    \[
    \alpha_k=\mathrm{median}(\{F_j:j\in N_{\mathrm{eval}}^{(k)}\}),\quad
    m_k=\max\left(\mathrm{median}(\{|F_j-\alpha_k|:j\in N_{\mathrm{eval}}^{(k)}\}),\epsilon\right),
    \]
    and define
    \[
    w_i =
    \sigma\left(\lambda\frac{F_i-\alpha_k}{m_k}\right)
    \frac{1}{1+o_i},
    \qquad
    \sigma(z)=\frac{1}{1+\exp(-z)}.
    \]
    The parent is sampled as
    \[
    p(i \mid k)=\frac{w_i}{\sum_{j\in N_{\mathrm{eval}}^{(k)}}w_j}.
    \]
    This favors high-scoring nodes while reducing the probability of repeatedly expanding nodes that have already produced many children. The island sampling promotes diversity across search directions, while the weighted parent sampling balances exploration and exploitation within each island.

    \item \textbf{Island-based diversity maintenance.}
    Diversity is maintained by evolving separate island subpopulations in parallel. This prevents all expansions from concentrating around a single high-performing branch. To share useful discoveries across the tree, nodes may occasionally migrate between islands, while island elites are protected from migration so that each island preserves its own best search direction.

    \item \textbf{Novelty-based rejection sampling.}
    After a new candidate is generated, novelty-based rejection sampling is applied before adding it to the tree. The mutable portion of the candidate is embedded and compared against existing nodes in the selected island using cosine similarity. If the maximum similarity exceeds a threshold $\eta$, the candidate is considered potentially redundant. In this case, an LLM novelty judge is queried to determine whether the candidate is meaningfully different; candidates judged insufficiently novel are rejected and resampled.

    \item \textbf{Inspiration sampling.}
    In addition to the main parent, the prompt includes inspiration nodes sampled from the tree. Following ShinkaEvolve~\cite{lange2025shinkaevolve}, these inspirations are drawn from a mixture of high-performing nodes and randomly sampled nodes, using the selected island as the local sampling context. The high-performing inspirations expose the LLM to successful strategies, while the random inspirations encourage recombination with more diverse parts of the tree search.

    \item \textbf{Mutation-type sampling.}
    Each expansion samples one of three mutation types: targeted diff edits, full rewrites, or crossover. We use probabilities
    \[
    p_{\mathrm{diff}} = 0.6,\qquad
    p_{\mathrm{full}} = 0.3,\qquad
    p_{\mathrm{cross}} = 0.1.
    \]
    Diff edits make localized changes to the parent node, full rewrites allow larger structural changes while preserving immutable code regions, and crossover samples an additional node and asks the LLM to combine ideas from both nodes.
\end{itemize}




\section{Evaluation}
\label{app:evaluation}

\subsection{Task Descriptions}
\label{sec:tasks}

\begin{figure}[H]
    \centering
    \includegraphics[width=\textwidth]{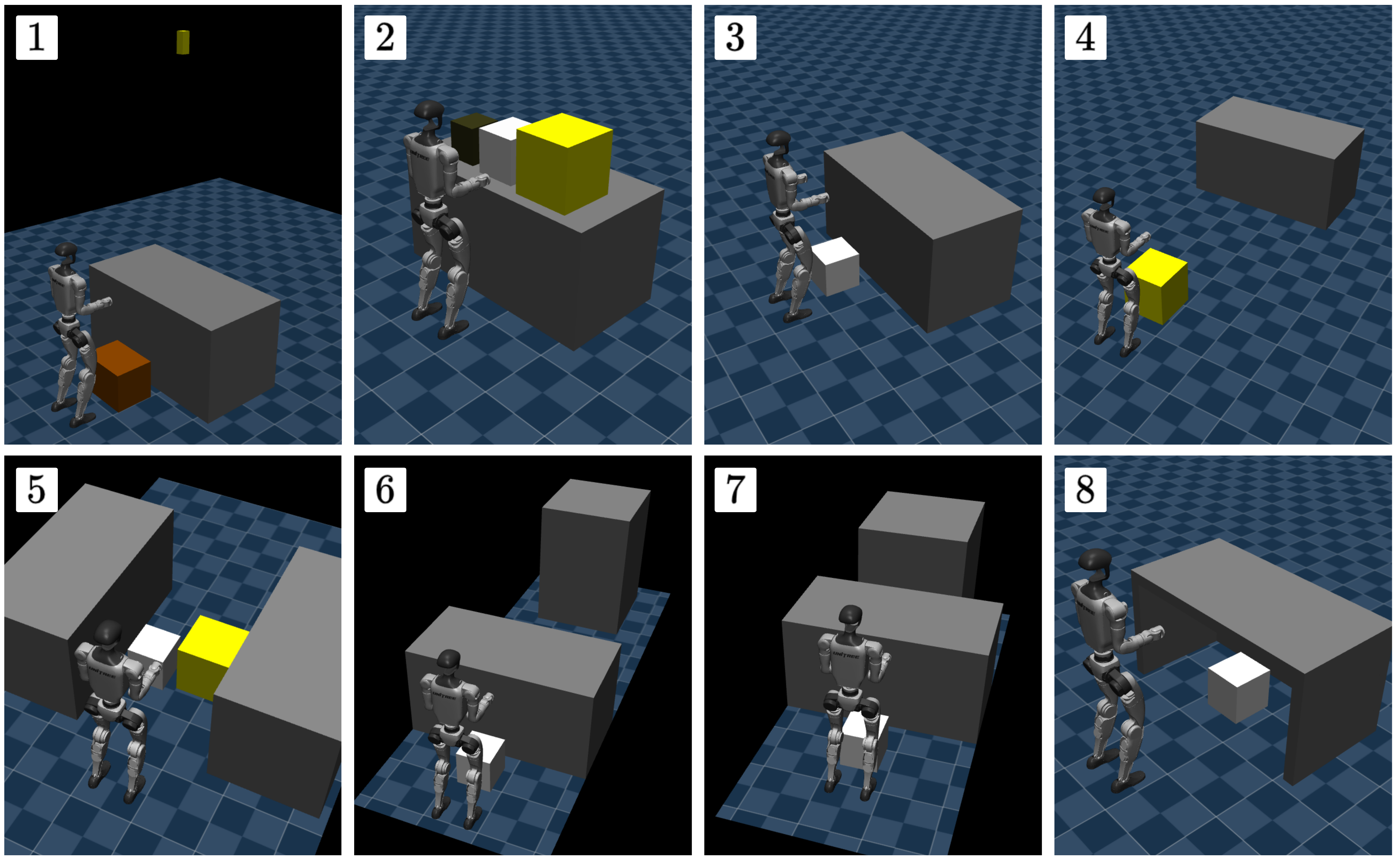}
    \caption{\textbf{Experiment scenes.} The eight loco-manipulation evaluation tasks, numbered as referenced in the text: (1) Banana, (2) Box Stacking, (3) Climb Table w/ Box, (4) Long-Dist.\ Pick \& Place, (5) Move Through Clutter, (6) Parkour Pick \& Place 1, (7) Parkour Pick \& Place 2, and (8) Under-Table Pick \& Place.}
    \label{fig:experiment_scenes}
\end{figure}

We describe below the eight loco-manipulation evaluation tasks. Figure~\ref{fig:experiment_scenes} shows the corresponding scenes.

\begin{itemize}
    \item \textbf{(1) Banana.} The robot must grasp a banana placed out of reach, which requires reasoning about stacking the object in the scene on top of the table and climbing on top of the object placed on the table to gain height.
    \item \textbf{(2) Box Stacking.} The robot must stack three boxes on top of each other to build the tallest possible stack, testing sequential manipulation and stability reasoning.
    \item \textbf{(3) Climb Table w/ Box.} This task tightly couples locomotion and manipulation: the robot must climb onto a table while moving a box with it throughout the maneuver.
    \item \textbf{(4) Long-Distance Pick \& Place.} The robot must transport a box and place it on a distant table, evaluating the ability to combine extended locomotion with manipulation.
    \item \textbf{(5) Move Through Clutter.} The robot must traverse a corridor obstructed by two boxes, probing obstacle-aware planning and interaction to reach the other side.
    \item \textbf{(6, 7) Parkour Pick \& Place 1 and 2.} These two variants pose increasingly difficult challenges. In the first, the robot must climb onto a table in order to place a box on an otherwise obstructed surface. In the second, more challenging variant, the robot must climb onto and back down from an intermediate table while carrying the box before placing it on the floor.
    \item \textbf{(8) Under-Table Pick \& Place.} The robot must reach into a confined space beneath a table, demanding low-clearance whole-body postures that go beyond standard standing manipulation.
\end{itemize}

\subsection{Real-World Deployment Snapshots}
\label{sec:real_world_deployment}

\begin{figure}[H]
    \centering
    \includegraphics[width=\textwidth]{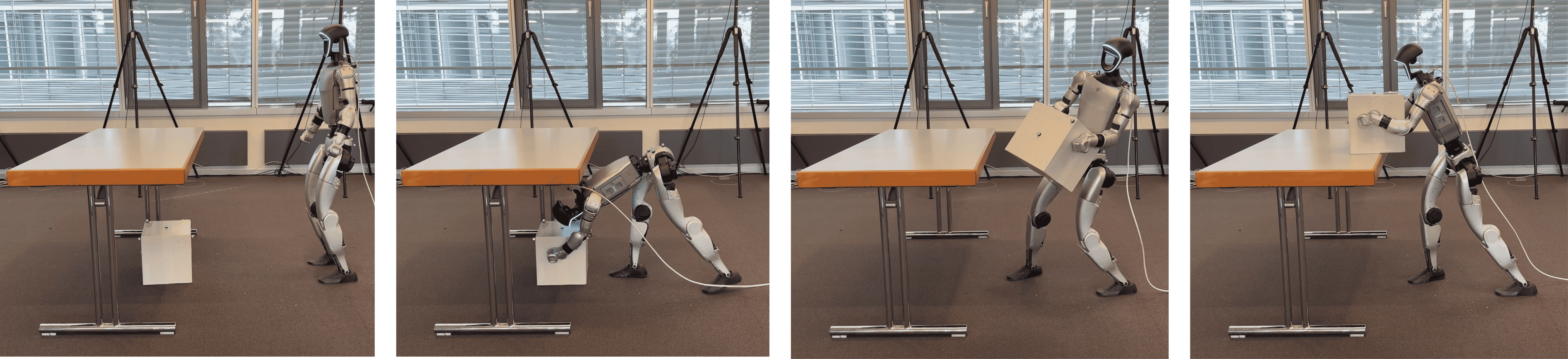}
    \caption{\textbf{Under-Table Pick \& Place.} Real-world snapshots of the humanoid robot reaching into a confined space beneath a table to retrieve the box, demonstrating the low-clearance whole-body posture discovered by MotionDisco.}
    \label{fig:real_world_under_table}
\end{figure}

\begin{figure}[H]
    \centering
    \includegraphics[width=\textwidth]{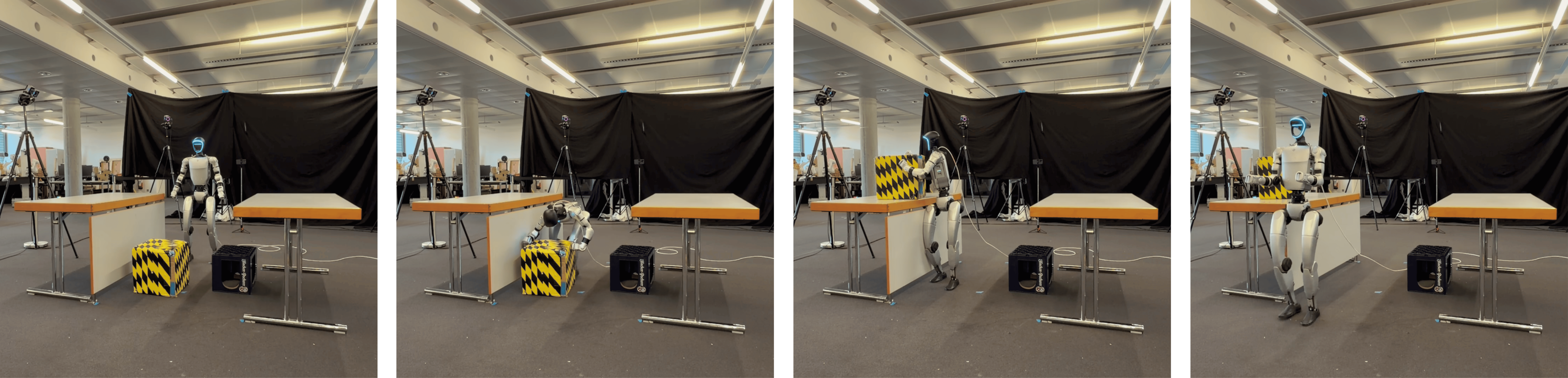}
    \caption{\textbf{Move Through Clutter.} Real-world snapshots of the humanoid robot traversing a corridor obstructed by two boxes, interacting with the obstacles along the path to make way for passage.}
    \label{fig:real_world_move_clutter}
\end{figure}

\begin{figure}[H]
    \centering
    \includegraphics[width=\textwidth]{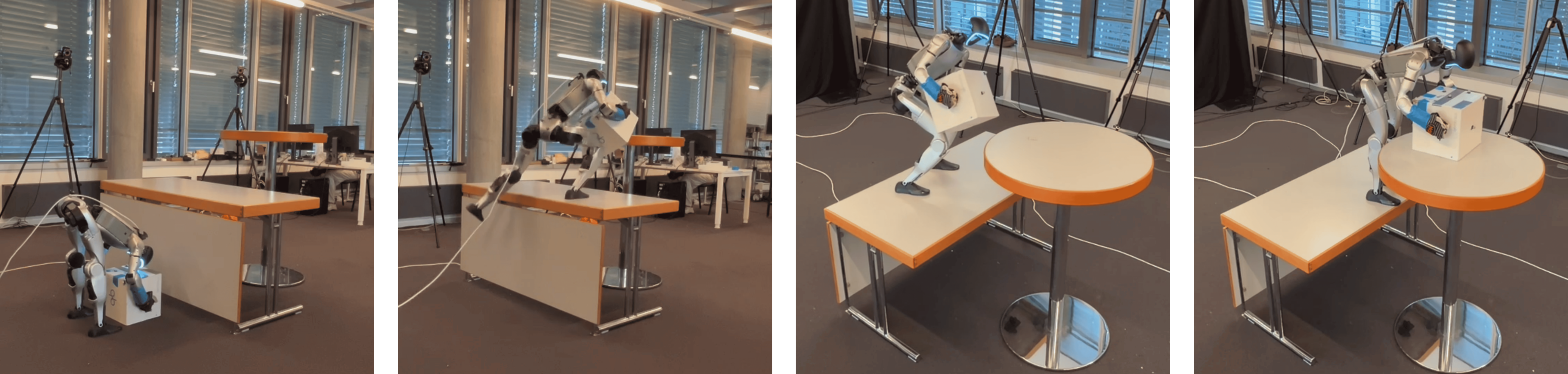}
    \caption{\textbf{Parkour Pick \& Place 2.} Real-world snapshots of the humanoid robot climbing onto the table using its feet, while carrying the box, before placing it on a distant table after taking some steps.}
    \label{fig:real_world_parkour_2}
\end{figure}

\begin{figure}[H]
    \centering
    \includegraphics[width=\textwidth]{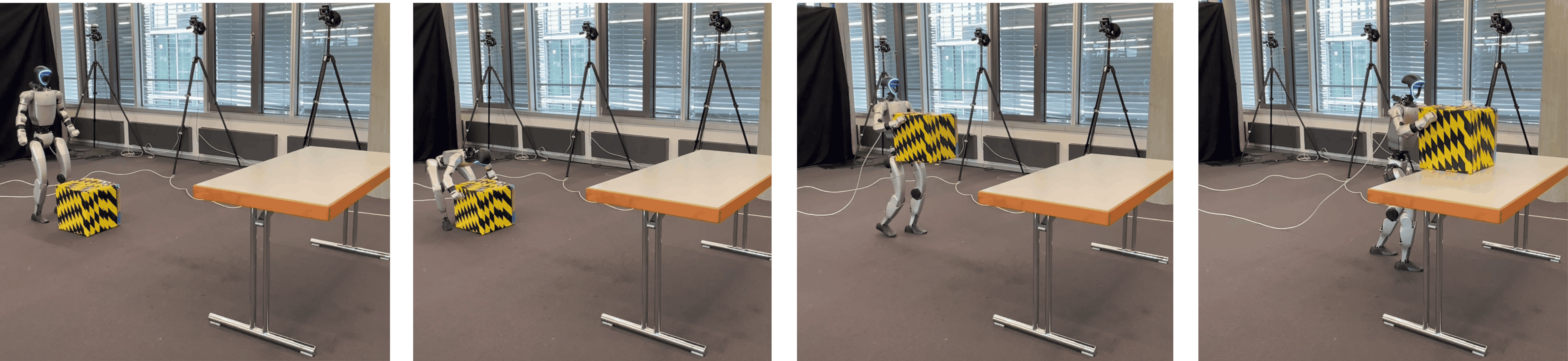}
    \caption{\textbf{Long-Distance Pick \& Place.} Real-world snapshots of the humanoid robot transporting a box and placing it on a distant table, combining extended locomotion with manipulation.}
    \label{fig:real_world_long_dist}
\end{figure}

\begin{figure}[H]
    \centering
    \includegraphics[width=\textwidth]{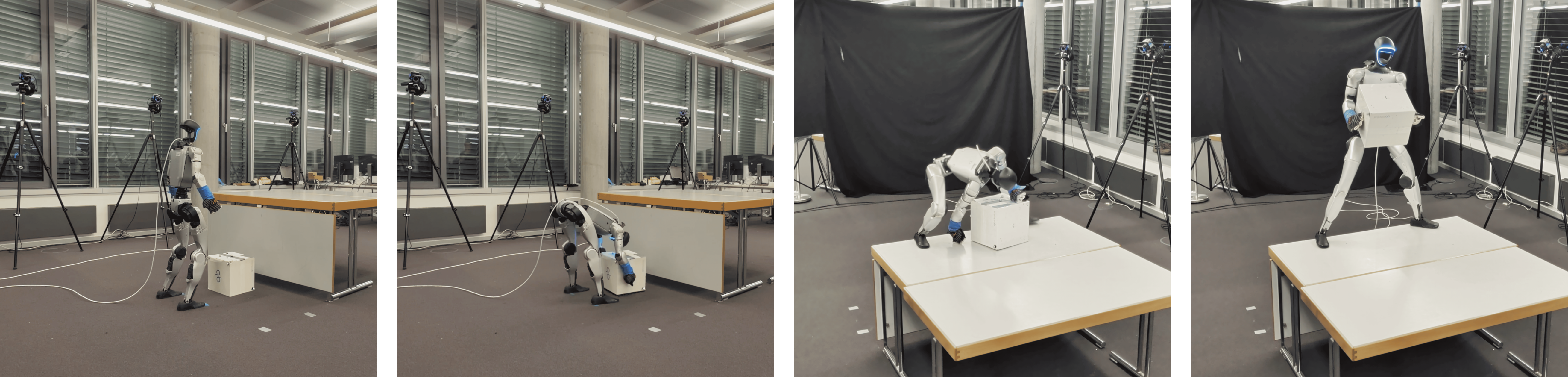}
    \caption{\textbf{Climb Table w/ Box.} Real-world snapshots of the humanoid robot climbing onto a table while moving a box, using its hands as support on the table, and lifting up the box.}
    \label{fig:real_world_climb_table}
\end{figure}

\end{document}